\documentclass{article}

\usepackage{arxiv}
\usepackage{graphicx}
\usepackage{comment}
\usepackage{amsmath,amssymb} 
\usepackage{color}

\def\bsq#1{
	\lq{#1}\rq
}
\usepackage{cite}

\usepackage{xcolor}

\usepackage{amsmath}
\usepackage{algorithm}
\usepackage[noend]{algpseudocode}

\newcommand\blfootnote[1]{%
	\begingroup
	\renewcommand\thefootnote{}\footnote{#1}%
	\addtocounter{footnote}{-1}%
	\endgroup
}

\algnewcommand{\Initialize}[1]{%
	\State \textbf{Initialize:}
	\State \hspace*{\algorithmicindent}\parbox[t]{0.8\linewidth}{\raggedright #1}
}
\usepackage[hang,flushmargin]{footmisc}

\usepackage{amsfonts} 
\DeclareMathOperator*{\argmax}{arg\,max}
\usepackage{booktabs} 
\usepackage{graphicx}
\usepackage{stmaryrd} 
\usepackage[tight,footnotesize]{subfigure}

\usepackage{adjustbox}

\usepackage[para]{threeparttable}
\usepackage{array}

\usepackage{authblk}

\title{Regularization with Latent Space Virtual Adversarial Training}

\author[1,2,5]{Genki Osada}
\author[1,3]{Budrul Ahsan}
\author[4]{Revoti Prasad Bora}
\author[2]{Takashi Nishide}
\affil[1]{Philips Co-Creation Center, Japan}
\affil[2]{University of Tsukuba, Japan}
\affil[3]{The Tokyo Foundation for Policy Research, Japan}
\affil[4]{Lowe's Services India Pvt. Ltd.}
\affil[5]{I Dragon Corporation, Japan}

\begin{document}
\maketitle
\begin{abstract}
	Virtual Adversarial Training (VAT) has shown impressive results among recently developed regularization methods called consistency regularization.
	VAT utilizes adversarial samples, generated by injecting perturbation in the input space, for training and thereby enhances the generalization ability of a classifier.
	However, such adversarial samples can be generated only within a very small area around the input data point, which limits the adversarial effectiveness of such samples.
	To address this problem we propose LVAT (Latent space VAT), which injects perturbation in the latent space instead of the input space.
	LVAT can generate adversarial samples flexibly, resulting in more adverse effect and thus more effective regularization.
	The latent space is built by a generative model, and in this paper we examine two different type of models: variational auto-encoder and normalizing flow, specifically Glow.
	We evaluated the performance of our method in both supervised and semi-supervised learning scenarios for an image classification task using SVHN and CIFAR-10 datasets.
	In our evaluation, we found that our method outperforms VAT and other state-of-the-art methods.
	\blfootnote{Accepted at ECCV 2020 (Oral).}
\end{abstract}

\section{Introduction}
One of the goals of training machine learning models is to avoid overfitting. To address overfitting, we use various regularization techniques.
Recently \textit{consistency regularization} has shown remarkable results to overcome overfitting problems in deep neural networks.
Consistency regularization is also known as perturbation-based methods, and its basic strategy is to perturb inputs during the learning process and force the model to be robust against them. Perturbation is defined as randomizing operations such as dropout, introducing Gaussian noise, and data augmentation.
Consistency regularization is achieved by introducing a regularization term, called consistency cost.
Consistency cost is designed to penalize the discrepancy between the model outputs, with and without perturbations.
Thus consistency regularization can work without class labels,
which leads to two attractive features:
1) it can extend most supervised learning methods to semi-supervised learning methods, which enables us to use unlabeled data for the model training, and 2) like the techniques such as dropout and data augmentation, it can be applied to most existing models without modifying the model architecture.

Among the consistency regularization methods, Virtual Adversarial Training (VAT) \cite{miyato2015distributional, miyato2018virtual} has shown promising results.
The way of generating perturbation in VAT is different from other consistency regularization methods.
Perturbation in VAT is not random, instead deliberately generated towards a direction that causes the most adverse effects on the model output.
However, perturbations in VAT can work only a very small area around the input data points.
It is because those perturbations are generated based on the local sensitivity, i.e., the gradients of the model outputs w.r.t. the tiny shift in the input space (Section \ref{sec:problem}).
In addition, the magnitude of perturbations is always fixed at the value specified by the hyperparameter, and is constant regardless of the input image.
We focus on the fact that such \textit{local constraint} limits the offensive power of adversarial perturbations, and thus hinders the effectiveness of VAT as a consistency regularization.
Based on this, we aim to overcome this limitation and develop a method to generate perturbation more flexibly, which would lead to better generalization.

In this paper, we propose VAT based consistency regularization that utilizes the latent space.
Our method computes the perturbation in the latent space, instead of the input space.
First, the input data is mapped to the latent space, and we compute perturbation and inject it into the point in the latent space.
Then, adversarial examples are generated by mapping such perturbed latent point back to the point in the input space, and therefore, there is no constraint which hinders VAT effectiveness.
To map to and from the latent space, we use a generative model, and in this paper we examine two models: variational auto-encoder (VAE) \cite{kingma2013auto} and normalizing flow, specifically, Glow \cite{NIPS2018_8224}.
We used SVHN and CIFAR-10 datasets, which are common benchmarking datasets, to demonstrate that our method improves VAT and outperforms other state-of-the-art methods.

\paragraph{Notation.}
We consider a classification task.
Suppose we have $K$ classes, taking an input ${\bf x}_{i} \in \mathcal{X}$, we want to predict the class label $y_{i} \in \{1,2,\ldots,K\}$\footnote{We write scalars and vectors by non-bold and bold letters, respectively.},
where $\mathcal{X}$ is a sample space with data points.
Our purpose is to learn $K$-class classifier model $f: \mathcal{X} \mapsto \mathbb{R}^{K}$ parameterized by $\theta$, where $\mathbb{R}$ denotes real numbers, the $k$-th element $f({\bf x})_{k}$ is called logit for class $k$,
and $y_{\text{pred}} = \argmax_{k} f({\bf x})_{k} \in \{1,2,\ldots,K\}$ corresponds to the model prediction of the class.
Also, labeled dataset is denoted by $D_{\text{l}} = \{({\bf x}_{i}, y_{i})\}_{i=1}^{N_{\text{l}}}$ with ${N_{\text{l}}}$ samples and unlabeled dataset is denoted by $D_{\text{u}} = \{{\bf x}_{i}\}_{i=1}^{N_{\text{u}}}$ with ${N_{\text{u}}}$ samples.

\section{Related Work}
\label{sec:relatedwork}
We focus mainly on state-of-the-art methods that are related to our approach.
These can be broadly categorized as follows: consistency regularization, graph-based methods, and GAN-based methods.

\paragraph{Consistency Regularization.}
The VAT and our proposed method LVAT belong to this category.
The assumption underlying these methods is \textit{local consistency} \cite{NIPS2003_2506}: nearby points in the input space are likely to have the same output.
In general, the model predictions for the points near the decision boundaries are sensitive to perturbations and prone to be misclassified by being perturbed.
To mitigate this sensitivity, this type of method employs the regularizing loss function, called consistency cost, which aims to train the model so that its outputs would be consistent for the inputs both with and without perturbation.
Because the consistency cost becomes large at the points near the decision boundaries,
the regularizing effect works so that the decision boundary would be kept far away from such points, which leads to better generalization in testing time.
As the simplest case, the $\Pi$-Model \cite{laine2016temporal} employs the following consistency cost $R({\bf x})$:
\begin{eqnarray} 
	R({\bf x}) := \lVert f(\tilde{{\bf x}}_{1}, \theta) - f(\tilde{{\bf x}}_{2}, \theta)  \rVert_{2}^{2} \\
	\tilde{{\bf x}}_{1} \sim \text{Perturb}({\bf x}), \ \tilde{{\bf x}}_{2} \sim \text{Perturb}({\bf x})
\end{eqnarray}
where $\lVert \cdot \rVert_{2}$ denotes $L_{2}$ norm, and \text{Perturb($\cdot$)} is a function that applies a stochastic deformation (i.e., data augmentation), random noise addition, and dropout, thus outputting different $\tilde{{\bf x}}_{i}$ each time.

Dropout, Gaussian noise, and randomized data augmentation have been chosen as perturbations in \cite{NIPS2017_6719, laine2016temporal, NIPS2016_6333, park2018adversarial, athiwaratkun2018there, ijcai2019-504, NIPS2019_8749}.
\cite{xie2019unsupervised} has shown that the latest data augmentation techniques, AutoAugment and Cutout, were quite effective to use as perturbations.
Although in the $\Pi$-Model, \text{Perturb($\cdot$)} is applied to both of ${\bf x}$ in $R({\bf x})$, \text{Perturb($\cdot$)} is applied only to one ${\bf x}$ in $R({\bf x})$ in VAT and LVAT.
Furthermore, perturbations used in VAT and LVAT are not random but are carefully computed, as we describe it in the next section.

\paragraph{Graph-based Methods.}
In contrast to the above consistency regularization methods, graph-based methods assume \textit{global consistency} \cite{NIPS2003_2506}: all samples that map to the same class label should belong to a single cluster.
\cite{pmlr-v80-kamnitsas18a} proposed a method that captures a structure of samples within a mini-batch by means of label propagation,
and then forces samples belonging to the same class to form compact clusters in the feature space.
Smooth Neighbors on Teacher Graphs (SNTG) \cite{Luo_2018_CVPR} computes unsupervised loss function for each mini-batch, in which attraction force between samples belonging to the same class and repulsion force between samples belonging to the different classes are realized.
Consistency regularizations focus on the sensitivity of each data point to perturbations,
whereas graph-based methods regularize a whole structure of data points within a mini-batch.
Graph-based methods and consistency regularizations are not mutually exclusive, but rather complementary.
These two could be implemented at the same time, and in fact \cite{Luo_2018_CVPR} reported that
combining with SNTG steadily improved the performance of all consistency regularization methods in their experiments.

\paragraph{GAN-based Methods.}
Several works have utilized the samples generated in GAN framework as another kind of perturbation injection \cite{springenberg2015unsupervised, salimans2016improved, NIPS2017_6997, dumoulin2016adversarially}.
Among them, BadGAN \cite{NIPS2017_7229} presented impressive performance.
As opposed to the usual GAN framework, the generator in BadGAN generates unrealistic samples,
which can be viewed as a data augmentation targeting the lower density regions in the given data distribution.
With such data augmentation, BadGAN aims to draw better decision boundaries, and thus their objective is similar to VAT and our proposed method.
We compare our method with these methods in result section.

\paragraph{Adversarial Examples for Generative Models.}
We finally note that, in the filed of adversarial machine learning, there are studies in regard to the latent space of generative models.
However, their objectives are not regularization like ours.
\cite{cao2017mitigating} studied attacking the generative models,  and \cite{NIPS2018_7394} derived a fundamental upper bound on robustness against adversarial perturbations.
Our goal is to achieve better consistency regularization, and to this end, we aim to generate more effective adversarial examples by utilizing the latent space.

\section{Preliminary}

\subsection{Virtual Adversarial Training}
\label{sec:vat}

Perturbations in VAT are deliberately generated so that its direction could cause the most adverse effects on the model outputs, i.e., classification predictions.
Formally, letting
\begin{eqnarray}
	R({\bf x}, {\bf r}) := \text{KL}( f({\bf x}, \theta) \parallel f({\bf x}+{\bf r}, \theta)), 
\end{eqnarray}
adversarial perturbation ${\bf r}_{\text{vat}}$ is defined as
\begin{eqnarray}
	{\bf r}_{\text{vat}} := \argmax_{\bf r} \{ R({\bf x}, {\bf r}); \lVert {\bf r} \rVert_{2} \leq \epsilon_{\text{vat}}\}
	\label{eq:rvat}
\end{eqnarray}
where KL($p \parallel q$) denotes Kullback-Leibler (KL) divergence between distributions $p$ and $q$,
and $\epsilon_{\text{vat}}$ is a hyper-parameter to decide the magnitude of ${\bf r}_{\text{vat}}$\footnote{We use the suffix of \bsq{vat} to distinguish from the symbols that will be used later in the description of our proposed method.}.
Once ${\bf r}_{\text{vat}}$ is computed, the consistency cost is given as:
\begin{eqnarray}
	\label{eq:loss_vat}
	L_{\text{vat}} &:=& R({\bf x}, {\bf r}_{\text{vat}})  \nonumber \\
	&=& \text{KL}( f({\bf x}, \theta) \parallel f({\bf x}+{\bf r}_{\text{vat}}, \theta)).
\end{eqnarray}
This regularizing cost encourages the classifier to be trained so that it outputs consistent predictions for the clean input ${\bf x}$ and the adversarially perturbed input ${\bf x}_{\text{adv}} = {\bf x}+{\bf r}_{\text{vat}}$.

Eq.\ (\ref{eq:rvat}) can be rewritten as ${\bf r}_{\text{vat}} = \epsilon_{\text{vat}} {\bf u}$, where ${\bf u}$ is a unit vector in the same
direction as ${\bf r}_{\text{vat}}$ and the maximum magnitude of ${\bf r}_{\text{vat}}$ is given by $\epsilon_{\text{vat}}$.
To calculate ${\bf u}$, \cite{miyato2015distributional} has presented following fast approximation method.
Under the assumption that $f({\bf x}, \theta)$ is twice differentiable with respect to $\theta$, the second-order Taylor expansion around the point of ${\bf r} = 0$ yields
\begin{eqnarray}
	\label{eq:Taylor}
	R({\bf x}, {\bf r}) &\approx& R({\bf x}, 0) + R'({\bf x}, 0) {\bf r} + \frac{1 }{2} {\bf r}^{T} H {\bf r} \\
	&=& \frac{1}{2} {\bf r}^{T} H {\bf r}
\end{eqnarray}
where $H$ is the Hessian matrix given by $H := R''({\bf x}, 0)$, and $R({\bf x}, 0)$ and $R'({\bf x}, 0)$ in the first line are zeros since KL($p \parallel q$) takes the minimal value zero when $p = q$, i.e., ${\bf r} = 0$.
Thus, taking the eigenvector of $H$ which has the largest eigenvalue is required to solve Eq.\ (\ref{eq:rvat}).
To reduce the computational cost, a finite difference power method is introduced.
Given a random unit vector ${\bf d}$, the iterative calculation of ${\bf d} \leftarrow \overline{H {\bf d}} $
where $\overline{H {\bf d}} := H{\bf d} / \lVert H{\bf d} \rVert_{2} $, makes the ${\bf d}$ converge to ${\bf u}$.
With a small constant $\xi$, finite difference approximation follows as
\begin{eqnarray} 
	H &\approx& \left( R'({\bf x}, 0 + \xi{\bf d}) - R'({\bf x}, 0) \right) / \xi{\bf d} \\
	Hd &=& R'({\bf x}, \xi{\bf d}) / \xi
\end{eqnarray}
where we use the fact that $R'({\bf x}, 0) = 0$.
Then the repeated application of ${\bf d} \leftarrow \overline{R'({\bf x}, \xi{\bf d})}$ yields ${\bf u}$.
\cite{miyato2015distributional} reported that sufficient result was reached by only one iteration.
As a result, with a given $\epsilon_{\text{vat}}$, Eq.\ (\ref{eq:rvat}) can be computed as:
\begin{eqnarray}
	\label{eq:rvat_final}
	{\bf r}_{\text{vat}} = \epsilon_{\text{vat}} \overline{R'({\bf x}, \xi{\bf d})}.
\end{eqnarray}
The pseudo-code describing the computation of $L_{\text{vat}}$ defined in Eq.\ (\ref{eq:loss_vat}) is shown in Algorithm \ref{algo:vat}, which will be helpful to clarify in which part our proposed method differs from the VAT.

\subsection{Generative Model}
\label{sec:generativemodel}
To map to and from the latent space, we use a generative model.
In this paper, we examine two types of model, VAE and normalizing flow.
The VAE is approximate inference and the dimensionality of the latent space is usually much smaller than that of the input space.
On the other hand, the normalizing flow is exact inference and the dimensionality of the latent space is kept equal to that of the input space, i.e., lossless conversion.
We will refer to these two methods collectively as \textit{generative model}.

\paragraph{Variational Auto-Encoder}
The Variational Auto-Encoder (VAE) consists of two networks: the encoder (\text{Enc}) that maps a data sample ${\bf x}$ to ${\bf z}$ in latent space,
and the decoder (\text{Dec}) that maps ${\bf z}$ back to a point $\hat{{\bf x}}$ in the input space as:
\begin{eqnarray} 
	{\bf z} \sim \text{Enc}({\bf x}) = q({\bf z}|{\bf x}), \  \hat{{\bf x}} \sim \text{Dec}({\bf z}) = p({\bf x}|{\bf z}) .
\end{eqnarray}
The VAE regularizes the encoder by imposing a prior over the latent distribution $p({\bf z})$.
Typically $p({\bf z})$ is set as a standard normal distribution $\mathcal{N}(0, \textbf{I})$.
The VAE loss is:
\begin{eqnarray} 
	\label{eq:vae}
	L_{\text{vae}} = - \mathbb{E}_{q({\bf z}|{\bf x})} \left[ \text{log} \ \frac{p({\bf x}|{\bf z}) p({\bf z})}{q({\bf z}|{\bf x})} \right]
\end{eqnarray}
and it can be written as the sum of the following two terms:
the expectation of negative log likelihood, i.e., the reconstruction error,
$\mathbb{E}_{q({\bf z}|{\bf x})} \left[ \text{log} \ p({\bf x}|{\bf z}) \right]$,
and a prior regularization term,
$\text{KL} ( q({\bf z}|{\bf x}) \parallel p({\bf z}))$.

\paragraph{Normalizing Flow}
Suppose $g(\cdot)$ is an invertible function and let ${\bf h}_0$ and ${\bf h}_1$ be random variables of equal dimensionality.
Under the change of variables rule, transformation ${\bf h}_1 = g ({\bf h}_0)$ can be written as the change in the probability density function:
$p({\bf h}_0) = p({\bf h}_1) | \text{det}(d {\bf h}_1 / d{\bf h}_0) |$.
Stacking this transformation $L$-times as ${\bf h}_1, {\bf h}_2, \ldots, {\bf h}_L$ yields:
\begin{eqnarray} 
	p({\bf h}_0) = p({\bf h}_L)  \prod_{i=1}^{L} \ | \text{det}(d {\bf h}_i / d{\bf h}_{i-1}) |,
\end{eqnarray}
and taking the logarithm results in:
\begin{eqnarray}
	\label{eq:nf}
	\text{log} \ p({\bf x}) = \text{log} \ p({\bf z}) + \sum_{i=1}^{L} \ \text{log} \ | \text{det}(d {\bf h}_i / d{\bf h}_{i-1}) |
\end{eqnarray}
where we define ${\bf h}_0 := {\bf x}$ and ${\bf h}_L := {\bf z}$.
Such a series of transformations can gradually transform $p({\bf x})$ into a target distribution $p({\bf z})$ of any form.
Setting $p({\bf z}) = \mathcal{N}(0, \textbf{I})$ is especially called a normalizing flow \cite{pmlr-v37-rezende15}.

As Eq.\ (\ref{eq:nf}) is the form of log-likelihood, the learning objective is to maximize $\mathbb{E}_{p({\bf x})} \text{log} \ p({\bf x})$ by optimizing $g(\cdot)$.
The function $g(\cdot)$ must be designed to have the tractability to compute its inverse and the determinant of Jacobian matrix $| \text{det}(d {\bf h}_i / d{\bf h}_{i-1}) |$ in Eq.\ (\ref{eq:nf}), and several methods have been proposed in this regard.
\textit{Autoregressive} models \cite{NIPS2016_6581, NIPS2017_6828, pmlr-v80-huang18d} have a powerful expression but are computationally slow due to non-parallelization.
Thus, we use \textit{split coupling} models \cite{dinh2014nice, dinh2016density}, specifically, Glow \cite{NIPS2018_8224}.
For brevity, we refer the reader to \cite{NIPS2018_8224}.

Similarly to the case of VAE, we denote transformation  ${\bf x} \rightarrow {\bf z}$ by ${\bf z} = \text{Enc}({\bf x})$ and ${\bf z} \rightarrow {\bf x}$ by ${\bf x} = \text{Dec}({\bf z})$, and we call them just \text{Enc}() and \text{Dec}() as generic notations, for convenience.

\section{Motivation}
\label{sec:problem}
\subsection{Problem: Local Constraint of VAT}

The performance of VAT hinges upon the offensive power of the perturbed examples it generates.
To make the VAT a more effective regularizer, we want to find more adverse perturbation ${\bf r}_{\text{vat}}$, which is our goal.
The perturbation is obtained as ${\bf r}_{\text{vat}} = \epsilon_{\text{vat}} \overline{R'({\bf x}, \xi{\bf d})}$ as Eq.\ (\ref{eq:rvat_final}).
What the VAT algorithm tells us is only the most adverse \bsq{direction}, a unit vector $ \overline{R'({\bf x}, \xi{\bf d})}$, and to get ${\bf r}_{\text{vat}}$ we multiply it by $\epsilon_{\text{vat}}$, which is given as a hyper-parameter.
It thus appears that by giving the larger $\epsilon_{\text{vat}}$ we could generate more adverse ${\bf r}_{\text{vat}}$, but it is not the case.
It is because the direction $\overline{R'({\bf x}, \xi{\bf d})}$ is computed based on the local sensitivity, i.e., the gradients of the model output w.r.t. the tiny shift in the input space.
When computing the direction, the VAT algorithm implicitly assumes that $\lVert {\bf r}_{\text{vat}} \rVert_{2} = \epsilon_{\text{vat}}$ is very small.
More specifically, Eq.\ (\ref{eq:Taylor}) is taken under the assumption that ${\bf r}$ is very small such that Taylor expansion is applicable.
In other words, the direction $\overline{R'({\bf x}, \xi{\bf d})}$ is reliable only when $\epsilon_{\text{vat}}$ is small.
Indeed, it has been empirically confirmed in \cite{miyato2018virtual} that a large $\epsilon_{\text{vat}}$ causes performance degradation.
Also, the opposite can happen.
It is possible that a smaller perturbation than the given $\epsilon_{\text{vat}}$ would be more effective.
We can often find unnaturally high-intensity pixels on the adversarial images that VAT generates (as shown in Figs.\ \ref{fig:svhn_vat} and \ref{fig:cifar10_vat} later), and we consider that such artifacts indicate that $\epsilon_{\text{vat}}$ is too large.
As such, the magnitude of the perturbation in VAT is always fixed at $\epsilon_{\text{vat}}$ regardless of images.
We refer to this limitation as \textit{local constraint} of VAT.


\subsection{Our Approach}
\label{sec:approach}
\begin{figure}[t]
	\centering
	\includegraphics[width=0.4 \textwidth]{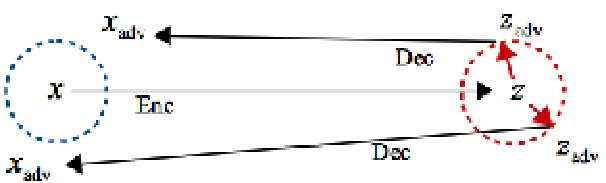}
	\caption{Illustration of our approach.
		Blue circle represents range of local constraint of VAT, whose radius is $\epsilon_{\text{vat}}$.
		A left-to-right arrow labeled \text{Enc} represents mapping from $\mathcal{X}$ to $\mathcal{Z}$.
		Red arrows represent ${\bf r}$ in $\mathcal{Z}$ and
		red circle represents $\epsilon_{\text{lvat}} = \lVert {\bf r} \rVert_{2}$ for our approach.
		Right-to-left arrows labeled \text{Dec} represent mapping from $\mathcal{Z}$ back to $\mathcal{X}$.
		Even though lengths (i.e., $\lVert  {\bf r}\rVert_{2}$) are equal for two red arrows,
		$\lVert {\bf x}_{\text{adv}} - {\bf x}\rVert_{2}$ depends on the direction of ${\bf r}$.
		We apply VAT to $\mathcal{Z}$ and select most effective ${\bf r}$ in $\mathcal{Z}$.
		${\bf x}_{\text{adv}}$ generated in this way is no longer constrained on blue circle, $\epsilon_{\text{vat}}$.}
	\label{fig:fig_epsilon}
\end{figure}
To make VAT a more effective regularizer, we aim to evade the local constraint of VAT and to make it possible to find adversarial perturbation more flexibly.
To this end, we introduce the latent space.
We first provide the intuition of the latent space built with the generative models, VAE and Normalizing Flow, introduced in Section \ref{sec:generativemodel}.
The latent space $\mathcal{Z}$ is trained such that every latent vector ${\bf z}$ has the one-to-one correspondence with ${\bf x} \in \mathcal{X}$
and that adjacent latent vectors ${\bf z}$ have corresponding ${\bf x}$ which are semantically continuous.
In other words, $\mathcal{Z}$ is the space where the data used to build $\mathcal{Z}$ are realigned so that they are semantically continuous.
Our approach takes advantage of this property of being semantically continuous.
Let us consider that we add a small random vector ${\bf r}$ to ${\bf z}$, which corresponds to a particular ${\bf x}$, and then we map that point ${\bf z} + {\bf r}$ back into the input space as ${\bf x}' = \text{Dec} (\text{Enc}({\bf x}) + {\bf r})$.
As long as $\lVert  {\bf r}\rVert_{2}$ is not much large,
the semantics of the obtained data ${\bf x}'$ will be preserved,
yet their visual appearance will deviate from the original input ${\bf x}$ without any constraint.
There is no constraint in terms of the distance between ${\bf x}'$ and ${\bf x}$ in $\mathcal{X}$.
To develop the intuition, we present an example of color changing.
Even when the background color of an image changes, its semantics remains unchanged, and thus the move in $\mathcal{Z}$ will be small.
However, the change in the pixel value, that is, the move that happened in $\mathcal{X}$ may well be huge.
This is our basic idea to avoid the local constraint, and instead of using the random vector ${\bf r}$, we compute the vector ${\bf r}_{\text{lvat}}$ in $\mathcal{Z}$ with VAT, which amounts to the most adversarial perturbation in $\mathcal{X}$.
To put it another way, our approach is to find the most adversarial perturbation in $\mathcal{X}$ with VAT in another space $\mathcal{Z}$, in which the data points are aligned differently from those in $\mathcal{X}$ and thus it can evade the constraint imposed in $\mathcal{X}$.
We note that although the magnitude of $\epsilon_{\text{lvat}} = \lVert {\bf r}_{\text{lvat}}\rVert_{2}$ has to be given as a hyper-parameter similarly to the original VAT, $\epsilon_{\text{lvat}}$ does not confine the distance $\lVert {\bf x}' - {\bf x}\rVert_{2}$ in $\mathcal{X}$.
We illustrate the overview of our approach in Fig.\ \ref{fig:fig_epsilon}.

In addition, performing VAT in $\mathcal{Z}$ brings us another advantage.
In the original VAT, conducting the grid-search on $\epsilon_{\text{vat}}$ is necessary because the optimal value for $\epsilon_{\text{vat}}$ heavily depends on the datasets, as shown in \cite{miyato2018virtual}.
By contrast, $\epsilon_{\text{lvat}}$ in our approach is much less susceptible to the difference in the datasets.
It is because regardless of the datasets, the latent space is constructed to form $\mathcal{N}(0, \textbf{I})$, and the scale of $\epsilon_{\text{lvat}}$ should not vary significantly from one dataset to another.
This is a strength of our approach in practical use.
In the next section we describe our methods.

\section{Method}
\label{sec:method}
\begin{figure}[t]

	\centering
	\includegraphics[width=0.75 \textwidth]{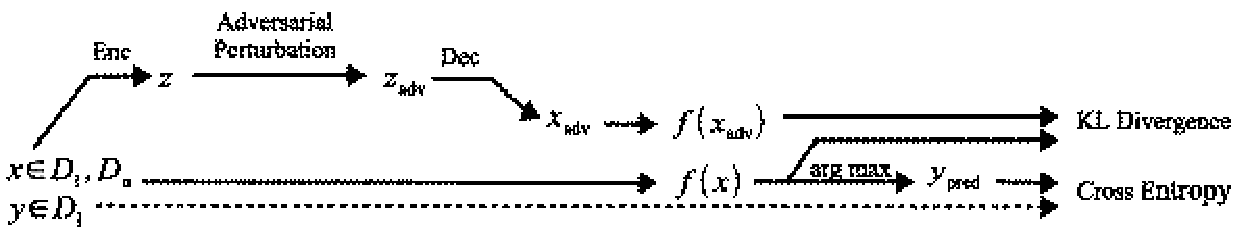}
	\caption{Overview of our method.
		Only during training, we place trained generative model, i.e., \text{Enc}() and \text{Dec}(), in front of classifier $f()$ being trained.
		While for ${\bf x} \in D_{\text{l}}, D_{\text{u}}$ classifier outputs $f({\bf x})$, predictive class label $y_{\text{pred}}$ is produced only for ${\bf x} \in D_{\text{l}}$.
		KL divergence corresponds to the consistency const $L_{\text{lvat}} $.}
	\label{fig:model}
\end{figure}

Our proposed method applies Eq.\ (\ref{eq:rvat}), i.e., Eq.\ (\ref{eq:rvat_final}), to the latent space.
It means that our method generates perturbations based on the gradients of the model outputs w.r.t. the shift in the latent space,
and therefore, the latent space in our method is required to be a continuous distribution.
Thus, vanilla Auto-Encoder and Denoising Auto-Encoder \cite{vincent2010stacked}, which do not construct the latent space as a continuous distribution, are out of our selection.
Instead, we choose two different types of generative models, VAE and Glow, and we build those models so that the latent space $p({\bf z})$ forms $\mathcal{N}(0, \textbf{I})$.
We call our proposed method \textit{LVAT} standing for Virtual Adversarial Training in the Latent space,
and we refer to LVAT using VAE and LVAT using Glow as \textit{LVAT-VAE} and \textit{LVAT-Glow}, respectively.
In Fig.\ \ref{fig:model}, we show the overview of LVAT.
We deploy the generative model in the fore stage of the classifier that we want to train.
During training, by mapping the input ${\bf x} \in D_{\text{l}}, D_{\text{u}}$ to the latent space by \text{Enc}(), the latent representation ${\bf z} = \text{Enc}({\bf x})$ is computed.
It is followed by applying Eq.\ (\ref{eq:rvat}) to ${\bf z}$ and computing the adversarial perturbation in the latent space, ${\bf r}_{\text{lvat}}$, and the adversarial latent representation ${\bf z}_{\text{adv}} = {\bf z} + {\bf r}_{\text{lvat}}$ is computed.
Then, by putting ${\bf z}_{\text{adv}}$ through \text{Dec}(), we obtain adversarial samples ${\bf x}_{\text{adv}} = \text{Dec}({\bf z}_{\text{adv}})$.
Here, we define
\begin{eqnarray} 
	R_{\text{lvat}}({\bf x}, {\bf r}) &:=& \text{KL}( f({\bf x}, \theta) \parallel f({\bf x}', \theta)) \\
	{\bf x}' &=& \text{Dec} (\text{Enc}({\bf x}) + {\bf r})
\end{eqnarray}
and the adversarial perturbation ${\bf r}_{\text{lvat}}$ and the consistency cost $L_{\text{lvat}}$ are defined as:
\begin{eqnarray}
	\label{eq:loss_lvat}
	{\bf r}_{\text{lvat}} &:=& \argmax_{\bf r} \{ R_{\text{lvat}}({\bf x}, {\bf r}) ; \lVert {\bf r} \rVert_{2} \leq \epsilon_{\text{lvat}}\}  \label{eq:r_lvat}\\
	{\bf x}_{\text{adv}} &:=& \text{Dec} (\text{Enc}({\bf x}) + {\bf r}_{\text{lvat}}) \\ 
	L_{\text{lvat}} &:=& R_{\text{lvat}}({\bf x}, {\bf r}_{\text{lvat}})   \\ 
	&=& \text{KL}( f({\bf x}, \theta) \parallel f({\bf x}_{\text{adv}}, \theta)).  
\end{eqnarray}
where $\epsilon_{\text{lvat}}$ is a hyper-parameter to decide the magnitude of ${\bf r}_{\text{lvat}}$.
The $\epsilon_{\text{vat}}$ in VAT gives the $L_{2}$ distance $\lVert {\bf x} - {\bf x}_{\text{adv}}\rVert_{2}$ in the input space,
whereas $\epsilon_{\text{lvat}}$ in LVAT gives the $L_{2}$ distance between $\lVert {\bf z} - {\bf z}_{\text{adv}}\rVert_{2}$ in the latent space.
The pseudo-code to obtain $L_{\text{lvat}}$ is shown in Algorithm \ref{algo:lvat}.

The full loss function $L$ is thus given by
\begin{eqnarray}
	L &=& L_{\text{sl}} (D_{\text{l}}, \theta) + \alpha L_{\text{usl}} (D_{\text{l}}, D_{\text{u}}, \theta) \label{eq:L}\\
	L_{\text{usl}} &=& \mathbb{E}_{{\bf x} \in D_{\text{l}}, D_{\text{u}}} [ L_{\text{lvat}} ] \nonumber
\end{eqnarray}
where $L_{\text{usl}}$ is the unsupervised loss, i.e., consistency cost,  $L_{\text{sl}}$ is a typical supervised loss (cross-entropy for our task), and $\alpha $ is a coefficient relative to the supervised cost.


\begin{algorithm}[t]
	\caption{Computation of consistency cost for VAT}\label{algo:vat}
	\begin{algorithmic}[1]
		\Require{$X$: random mini-batch from dataset}
		\Require{$f()$: classifier being trained}
		\Require{$\epsilon_{\text{vat}}$: magnitude of perturbation}
		\Require{$\xi$: very small constant, e.g., $1\mathrm{e}{-6}$}
		\Require{${\bf d}$: random unit vector of same shape of $X$}
		\Ensure{$L_{\text{vat}}$: consistency cost of VAT}
		\State ${\bf g} \gets \nabla_{{\bf d}} $  \Call{KL}{$ f(X) \parallel f(X + \xi {\bf d}$}
		\State ${\bf r}_{\text{vat}} \gets \epsilon_{\text{vat}} {\bf g} / \lVert {\bf g} \rVert_{2} $
		\State $L_{\text{vat}} \gets$ \Call{KL}{$ f(X) \parallel f(X + {\bf r}_{\text{vat}})$}
		\State \Return $L_{\text{vat}}$
	\end{algorithmic}
\end{algorithm}

\begin{algorithm}[t]
	\caption{Computation of consistency cost for LVAT}\label{algo:lvat}
	\begin{algorithmic}[1]
		\Require{$X$: random mini-batch from dataset}
		\Require{$f()$: classifier being trained}
		\Require{Enc() and Dec(): encoder and decoder of generative model}
		\Require{$\epsilon_{\text{lvat}}$: magnitude of perturbation}
		\Require{$\xi$: very small constant, e.g., $1\mathrm{e}{-6}$}
		\Require{${\bf d}$: random unit vector of same size as latent space}
		\Ensure{$L_{\text{lvat}}$: consistency cost of LVAT}
		\State $Z \gets$ Enc($X$)
		\State ${\bf g} \gets \nabla_{{\bf d}} $  \Call{KL}{$ f(X) \parallel f$(Dec($Z + \xi {\bf d}$))}
		\State ${\bf r}_{\text{lvat}} \gets \epsilon_{\text{lvat}} {\bf g} / \lVert {\bf g} \rVert_{2} $
		\State $L_{\text{lvat}} \gets$ \Call{KL}{$ f(X) \parallel f$(Dec($Z + {\bf r}_{\text{lvat}}$))}
		\State \Return $L_{\text{lvat}}$
	\end{algorithmic}
\end{algorithm}

\section{Experiments}
We evaluate our proposed method in an image classification task using SVHN and CIFAR-10 datasets.
Both supervised learning (SL) and semi-supervised learning (SSL) tests are conducted.
The experimental code\footnote{https://github.com/geosada/LVAT} was run with NVIDIA GeForce GTX 1070.

\subsection{Datasets}
The street view house numbers (SVHN) dataset consists of $32 \times 32$ pixel RGB images of real-world house numbers, having $10$ classes.
The CIFAR-10 dataset also consists of $32 \times 32$ pixel RGB images in $10$ different classes, \textit{airplanes, cars, birds, cats, deer, dogs, frogs, horses, ships}, and \textit{trucks}.
The numbers of training/test images are $73,257$/$26,032$ for SVHN and $50,000$/$10,000$ for CIFAR-10, respectively.

We also evaluate our method using augmented datasets.
We augmented data using random $2 \times 2$ translation for both datasets and horizontal flips only for CIFAR-10 same as previous study \cite{miyato2018virtual}.
These augmentations are dynamically applied for each mini-batch.
We denote the datasets with data augmentation by (w/ aug.), and our evaluation is conducted with four datasets, SVHN, SVHN (w/ aug.), CIFAR-10, and CIFAR-10 (w/ aug.).

In tests in SL, all labels in the training dataset are used, and the results are averaged over 3 runs.
In tests in SSL, $1,000$ and $4,000$ labeled data points are randomly sampled for SVHN and CIFAR-10, respectively.
To evaluate different combinations of labeled data in tests in SSL,
we prepared $5$ different datasets with $5$ different seeds for random sampling of labeled data points,
and the results are averaged over them.

\subsection{Classifier and Generative models}
The generative model can be modularized in our method, and thus we first train only the generative model for each dataset separately from the classifier.
Once we build the generative models, then, training the classifier with LVAT using the trained generative model follows.
We can use the same generative models throughout all the experiments, which benefits us as it reduces experiment time significantly, especially when we have to run the experiments many times (e.g., for grid-searching for hyper-parameters).
This can be viewed as a sort of curriculum strategy that is found for example in \cite{Li_2017_CVPR}.

\subsubsection{Model architectures.}
\begin{table*}[t]
	\caption{Architecture of classifier. BNorm stands for batch normalization \cite{pmlr-v37-ioffe15}. Slopes of all Leaky ReLU (lReLU) \cite{Maas13rectifiernonlinearities} are set to $0.1$.}
	\centering
		\begin{tabular}{ll}
			\toprule
			Input: $32 \times 32$ RGB image \ & \ 8: $2 \times 2$ max-pool, dropout 0.5 \\
			1: $3 \times 3$ conv. 128 same padding, BNorm, lReLU \ & \ 9: $3 \times 3$ conv. 512 valid padding, BNorm, lReLU\\
			2: $3 \times 3$ conv. 128 same padding, BNorm, lReLU \ & 10: $1 \times 1$ conv. 256 BNorm, lReLU\\
			3: $3 \times 3$ conv. 128 same padding, BNorm, lReLU \ & 11: $1 \times 1$ conv. 128 BNorm, lReLU  \\
			4: $2 \times 2$ max-pool, dropout 0.5 \ & 12: Global average pool $6 \times 6$ $\rightarrow$ \ $1 \times 1$\\
			5: $3 \times 3$ conv. 256 same padding, BNorm, lReLU \ & 13: Fully connected 128 $\rightarrow$ 10 \\
			6: $3 \times 3$ conv. 256 same padding, BNorm, lReLU \ & 14: BNorm (only for SVHN)\\
			7: $3 \times 3$ conv. 256 same padding, BNorm, lReLU \ & 15: Softmax\\
			\bottomrule
		\end{tabular}
	\label{tab:classifier}
\end{table*}
\begin{table*}[t]
	\caption{Architecture of encoder and decoder of VAE. Dimensionality of latent space is 128. BNorm stands for batch normalization. Slopes of Leaky ReLU (lReLU) are set to $0.1$.}
	\centering
		\begin{tabular}{ll}
			\toprule
			Enc & Dec \\
			\midrule
			Input: $32 \times 32 \times 3$ image & Input: 128-dimensional vector\\
			$2 \times 2$ conv. 128 valid padding, BNorm, ReLU \ & Fully connected 128 $\rightarrow$ 512 ($ 4 \times 4 \times 32$), lReLU \\
			$2 \times 2$ conv. 256 valid padding, BNorm, ReLU \ & $2 \times 2$ deconv. 512 same padding, BNorm, ReLU \\
			$2 \times 2$ conv. 512 valid padding, BNorm, tanh & $2 \times 2$ deconv. 256 same padding, BNorm, ReLU \\
			Fully connected 8192 $\rightarrow$ 128     & $2 \times 2$ deconv. 128 same padding, BNorm, ReLU \\
			& $1 \times 1$ conv. 128 valid padding, sigmoid \\
			\bottomrule
		\end{tabular}
	\label{tab:vae}
\end{table*}
The architecture of the classifier is the same as that of the previous works, and the detail is shown in Table \ref{tab:classifier}.
The architecture of the VAE is designed based on DCGAN \cite{radford2015unsupervised}, which is shown in Table \ref{tab:vae}.
In order to stabilize the training process, we apply a coefficient of 0.1 to the prior term, $\text{KL} ( q({\bf z}|{\bf x}) \parallel p({\bf z}))$ in Eq.\ (\ref{eq:vae}).
The architecture of the Glow mainly consists of two parameters: the depth of flow $K$ and the number of levels $L$.
The set of affine coupling and $1 \times 1$ convolution are performed $K$ times, followed by the factoring-out operation, and this sequence is repeated $L$ times.
We set $K=22$ and $L=3$, respectively\footnote{We implemented Glow model based on \cite{code_Glow}.}.
To avoid the gradient explosion in computing the loss during the training, the scaling factor of the affine coupling layer, $s$, is often restricted to avoid becoming too large.
One typical restriction is applying the sigmoid function to $s$ as in \cite{pmlr-v130-behrmann21a}, but we apply clipping as $\text{min}(|s|, 15)$.
Refer to \cite{NIPS2018_8224, dinh2016density} and our experimental code for more details.

\subsubsection{Hyper-parameters for classifier.}
\label{sec:param}
For the classifier with LVAT, we fixed the coefficient $\alpha = 1$ in Eq.\ (\ref{eq:L}), like the original VAT.
We used the Adam optimizer \cite{kingma2014adam} with the momentum parameters $\beta_{1} = 0.9$ and $\beta_{2} = 0.999$.
The initial learning rate is set to 0.001 and decays linearly with the last 16,000 updates, and $\beta_{1}$ is changed to 0.5 when the learning rate starts decaying.
The size of a mini-batch is 32 and 128 for $L_{\text{sl}}$ and $L_{\text{usl}}$, respectively for both datasets.
We trained each model with 48,000 and 200,000 updates for SVHN and CIFAR-10, respectively.
The best hyper-parameter $\epsilon_{\text{lvat}}$ in Eq.\ (\ref{eq:r_lvat}) was found through a grid search in the SSL setting.
For LVAT-VAE, $1.5$ and $1.0$ were selected for SVHN and CIFAR-10, respectively, from \{0.1, 0.25, 0.5, 0.75, 3.0, 4.0, ..., 15.0\}.
For LVAT-Glow, $1.0$ was selected for both SVHN and CIFAR-10 from \{0.5, 1.0, 1.5\}.

Also for a fair comparison, we conduct the test in SL for VAT, since the results for these were not reported in the original paper except for the one on CIFAR-10 (w/ aug.).
According to the code the original authors provide\footnote{https://github.com/takerum/vat\_tf},
we set $\epsilon_{\text{vat}}$ to $2.5, 3.5, 10.0$, and $8.0$ for SVHN, SVHN (w/ aug.), CIFAR-10, and CIFAR-10 (w/ aug.), respectively.
Although these values are provided for SSL test and we also attempted other values, it was found that the above ones were better.
Regarding $\epsilon_{\text{lvat}}$ and $\epsilon_{\text{vat}}$, we use these values for LVAT and VAT for all experiments unless otherwise noted.

\subsubsection{Training process of generative models.}
\label{sec:glow_and_vae}
\begin{figure}[t]
	\centering
	\subfigure[VAE]{
		\includegraphics[width=0.22 \textwidth]{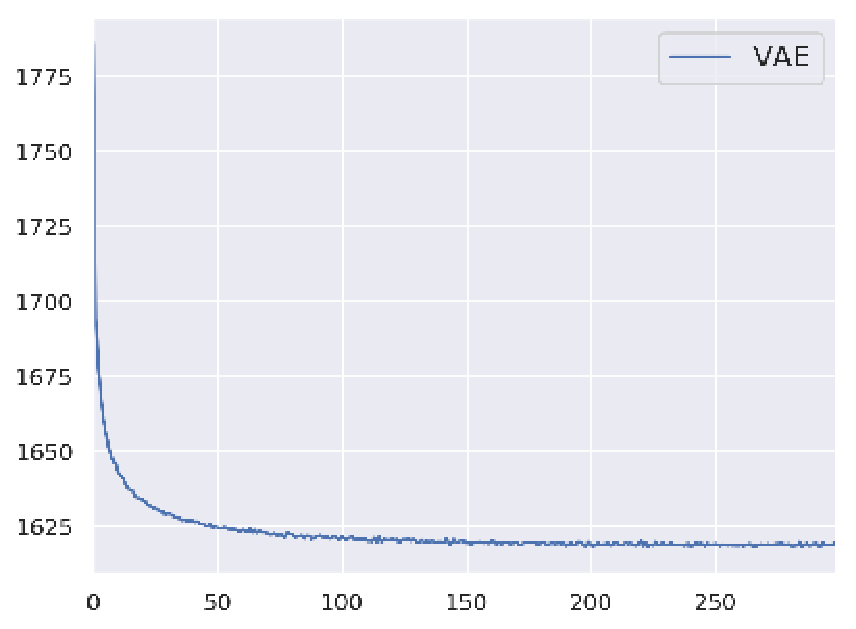}
	}
	\subfigure[Glow]{
		\includegraphics[width=0.22 \textwidth]{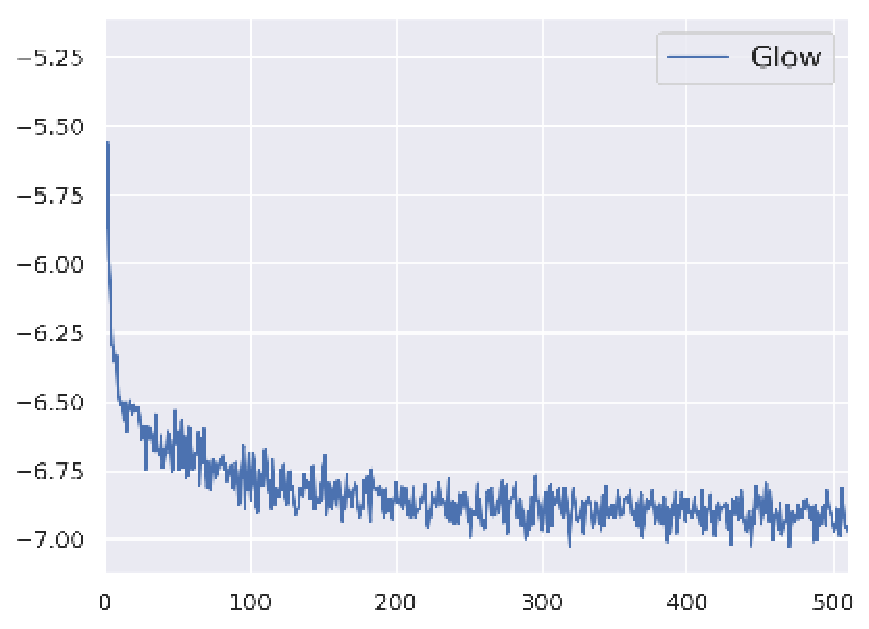}
	}
	\caption{Training loss during model training of generative models on CIFAR-10. Data augmentation were applied to both.
	}
	\label{fig:glow_and_vae}
\end{figure}
The VAE and Glow were trained using the same data that was used for training the classifier.
For the VAE, we also used the Adam optimizer with $\beta_{1} = 0.9$ and $\beta_{2} = 0.999$, with batch size $256$.
The learning rate starts with $0.001$ and exponentially decays with rate $0.97$ at every $2$ epochs after the first $80$ epochs, and we trained for $300$ epochs.
For the Glow, the learning rate starts with $0.0001$, and we trained $15600$ iterations for SVHN and CIFAR-10 and $25600$ iterations for SVHN (w/ aug.) and CIFAR-10 (w/ aug.).
The respective training processes on CIFAR-10 (w/ aug.) are shown in Fig.\ \ref{fig:glow_and_vae}.
For VAE, 1 epoch corresponds to one cycle of the entire training dataset, and the value of loss corresponds to Eq.\ (\ref{eq:vae}).
For Glow, 50 iterations are denoted as one epoch, and the loss is $- \text{log}_{\text{e}} p({\bf x})$, corresponding to Eq.\ (\ref{eq:nf}), which we normalized by the image size for implementation convenience.
To convert the units of loss value to \bsq{bits per dimension}, which is often used in the literature, we divide the value by 3, which is the number of RGB channels, and then divide it by $\text{log}_{\text{e}} 2$ for the conversion from nat to bit.
We can see that the training for both has converged, in particular the VAE has already converged well after about 100 epochs.
During the training of the classifier, the generative models are frozen and their parameters are not updated.

\subsection{Results}

\begin{table*}[t]
	\centering
	\caption{Error rates (\%) comparing to VAT and other methods. Results with data augmentation are denoted with (w/ aug.). SSL indicates semi-supervised learning, i.e., number of labeled data ${N_{\text{l}}}$ is 1,000 and 4,000 for SVHN and CIFAR-10, respectively. SL indicates supervised learning, i.e., all training data are used with label.}
	\begin{adjustbox}{max width=\textwidth}
		\begin{threeparttable}[t]

			\label{tab:vs_others}
			\begin{tabular}{lllllllll}
				\toprule
				& \multicolumn{2}{c}{SVHN} & \multicolumn{2}{c}{SVHN (w/ aug.)} & \multicolumn{2}{c}{CIFAR-10} & \multicolumn{2}{c}{CIFAR-10 (w/ aug.)}\\ \cmidrule(r){2-3} \cmidrule(r){4-5} \cmidrule(r){6-7} \cmidrule(r){8-9}
				Methods & SSL & SL & SSL & SL & SSL & SL & SSL	 & SL  \\
				\midrule
				\multicolumn{2}{l}{\bf{Consistency Regularization}} & & & & & & & \\
				\ \ Sajjadi {\em et al.} \cite{NIPS2016_6333} & - & - & - & 2.22 ($\pm$ 0.04) & - & 11.29 ($\pm$ 0.24) & - \\
				\ \ MT \cite{NIPS2017_6719} & 5.21 ($\pm$ 0.21) & 2.77 ($\pm$ 0.09) &  3.95 ($\pm$ 0.19) & 2.50 ($\pm$ 0.05) & 17.74 ($\pm$ 0.30) & 7.21 ($\pm$ 0.24) & 12.31 ($\pm$ 0.28) & 5.94 ($\pm$ 0.14) \\
				\ \ $\Pi$-Model \cite{laine2016temporal}& 5.43 ($\pm$ 0.25) & - & 4.82 ($\pm$ 0.17) & 2.54 ($\pm$ 0.04) & 16.55 ($\pm$ 0.29) & - & 12.36 ($\pm$ 0.31) & 5.56 ($\pm$ 0.10) \\
				\ \ TempEns \cite{laine2016temporal}& - & - & 4.42 ($\pm$ 0.16) & 2.74 ($\pm$ 0.06) & - & - & 12.16 ($\pm$ 0.24) & 5.60 ($\pm$ 0.14) \\
				\ \ VAT \cite{miyato2018virtual} & 5.77 ($\pm$ 0.32) & 2.34 ($\pm$ 0.05) \footnotemark[1]& 5.42 ($\pm$ 0.22) & 2.22 ($\pm$ 0.08) \footnotemark[1] & 16.92 ($\pm$ 0.45) \footnotemark[2] \footnotemark[3]& 8.175 \footnotemark[2]& 11.36 ($\pm$ 0.34) & $5.81 (\pm$ 0.02)\\
				\multicolumn{2}{l}{\bf{Graph-based Methods}} & & & & & & & \\
				\ \ LBA \footnotemark[4] \cite{Haeusser_2017_CVPR}& 9.25 ($\pm$ 0.65) & 3.61 ($\pm$ 0.10) & 9.25 ($\pm$ 0.65) & 3.61 ($\pm$ 0.10) & 19.33 ($\pm$ 0.51) & 8.46 ($\pm$ 0.18) & 19.33 ($\pm$ 0.51) & 8.46 ($\pm$ 0.18) \\
				\ \ CCLP \cite{pmlr-v80-kamnitsas18a}& 5.69 ($\pm$ 0.28) & 3.04 ($\pm$ 0.05) & - & - & 18.57 ($\pm$ 0.41) & 8.04 ($\pm$ 0.18) & - & - \\
				\multicolumn{2}{l}{\bf{GAN-based Methods}} & & & & & & & \\
				\ \ ALI \cite{dumoulin2016adversarially}& 7.42 ($\pm$ 0.65) & - & - & - & 17.99 ($\pm$ 1.62) & - & - & - \\
				\ \ CatGAN \cite{springenberg2015unsupervised} & - & - & - & - & 19.58 ($\pm$ 0.58) &9.38 & - & -\\
				\ \ TripleGAN \cite{NIPS2017_6997}& 5.77 ($\pm$ 0.17) & - & - & - & 16.99 ($\pm$ 0.36) & - & - & - \\
				\ \ ImprovedGAN \cite{salimans2016improved}& 8.11 ($\pm$ 1.30) & - & - & - & 18.63 ($\pm$ 2.32) & - & - & -  \\
				\ \ BadGAN \cite{NIPS2017_7229}& 4.25 ($\pm$ 0.03) & - & - & - & 14.41 ($\pm$ 0.30) & - & - & - \\
				\midrule
				LVAT-VAE (Ours) & 4.44 ($\pm$ 0.36) & {\bf 2.26} ($\pm$ 0.08) & 4.20 ($\pm$ 0.23)& {\bf 2.02} ($\pm$ 0.04)& 13.90 ($\pm$ 0.36) & 8.05 ($\pm$ 0.30) & 14.64 ($\pm$ 0.54) & 6.54 ($\pm$ 0.26) \\
				LVAT-Glow (Ours) &  {\bf 4.20} ($\pm$ 0.45) & {\bf 2.23} ($\pm$ 0.07) & {\bf 3.83} ($\pm$ 0.37)& {\bf 2.13} ($\pm$ 0.07)& {\bf 9.94} ($\pm$ 0.22)& {\bf 5.24} ($\pm$ 0.20) & {\bf 7.34} ($\pm$ 0.24) & {\bf 3.94} ($\pm$ 0.05) \\
				\bottomrule
			\end{tabular}
			\begin{tablenotes}\footnotesize
				\item [1] Results of our experiments with code \cite{miyato2018virtual} provided.
				\item [2] Results of our experiments with code \cite{miyato2018virtual} provided without ZCA.
				\item [3] Reported result in \cite{miyato2018virtual} is 14.87 ($\pm$ 0.38) with ZCA.
				\item [4] Results of re-implementation by \cite{pmlr-v80-kamnitsas18a}.
			\end{tablenotes}
		\end{threeparttable}
	\end{adjustbox}
\end{table*}

\begin{table*}[t]
	\centering
	\caption{Error rates (\%) comparing to combination methods. Notations are same as those in Table \ref{tab:vs_others}.}
	\begin{adjustbox}{max width=\textwidth}
		\begin{threeparttable}[t]

			\label{tab:vs_combi}
			\begin{tabular}{lllllllll}
				\toprule
				& \multicolumn{2}{c}{SVHN} & \multicolumn{2}{c}{SVHN (w/ aug.)} & \multicolumn{2}{c}{CIFAR-10} & \multicolumn{2}{c}{CIFAR-10 (w/ aug.)}\\ \cmidrule(r){2-3} \cmidrule(r){4-5} \cmidrule(r){6-7} \cmidrule(r){8-9}
				Methods & SSL & SL & SSL & SL & SSL & SL & SSL	 & SL  \\
				\midrule
				\multicolumn{2}{l}{\bf{Combination Methods}} & & & & & & & \\
				\ \ MT + SNTG \cite{Luo_2018_CVPR} & - & - &  3.86 ($\pm$ 0.27) & 2.42 ($\pm$ 0.06) & - & - & - & - \\
				\ \ MT + fast-SWA \cite{athiwaratkun2018there} & - & - &  - & - & - & - & 9.05 ($\pm$ 0.21) & 4.73 ($\pm$ 0.18) \\
				\ \ $\Pi$-Model + SNTG \cite{Luo_2018_CVPR} & 4.22 ($\pm$ 0.16) & - & {\bf 3.82} ($\pm$ 0.25) & 2.42 ($\pm$ 0.05) & 13.62 ($\pm$ 0.17) & - & 11.00 $(\pm$ 0.13) & 5.19 ($\pm$ 0.14) \\
				\ \ $\Pi$-Model + fast-SWA \cite{athiwaratkun2018there} & - & - &  - & - & - & - & 10.07 ($\pm$ 0.27) & 4.72 ($\pm$ 0.04) \\
				\ \ TempEns + SNTG \cite{Luo_2018_CVPR} & - & - & 3.98 ($\pm$ 0.21) & 2.44 ($\pm$ 0.03) & - & - & 10.93 ($\pm$ 0.14) & 5.20 ($\pm$ 0.14) \\
				\ \ VAT + Ent \cite{miyato2018virtual} & 4.28 ($\pm$ 0.10) & - & 3.86 ($\pm$ 0.11)  & - & 13.15 ($\pm$ 0.21) & - & 10.55 ($\pm$ 0.05) & - \\
				\ \ VAT + Ent + SNTG \cite{Luo_2018_CVPR} &{\bf  4.02} ($\pm$ 0.20) & - & 3.83 ($\pm$ 0.22) & - & 12.49 ($\pm$ 0.36) & - & 9.89 ($\pm$ 0.34)  & - \\
				\ \ VAT + Ent + fast-SWA \cite{athiwaratkun2018there} & - & - &  - & - & - & - & 10.97 & - \\
				\ \ VAT + LGA \cite{LGA}& 6.58 ($\pm$ 0.36) & - & - & - & 12.06 ($\pm$ 0.19) & - & - & - \\
				\midrule
				LVAT-VAE (Ours) & 4.44 ($\pm$ 0.36) & {\bf 2.26} ($\pm$ 0.08) & 4.20 ($\pm$ 0.23)& {\bf 2.02} ($\pm$ 0.04)& 13.90 ($\pm$ 0.36) & 8.05 ($\pm$ 0.30) & 14.64 ($\pm$ 0.54) & 6.54 ($\pm$ 0.26) \\
				LVAT-Glow (Ours) &  4.20 ($\pm$ 0.45) & {\bf 2.23} ($\pm$ 0.07) & 3.83 ($\pm$ 0.37)& {\bf 2.13} ($\pm$ 0.07)& {\bf 9.94} ($\pm$ 0.22)& {\bf 5.24} ($\pm$ 0.20) & {\bf 7.34} ($\pm$ 0.24) & {\bf 3.94} ($\pm$ 0.05) \\
				\bottomrule
			\end{tabular}
		\end{threeparttable}
	\end{adjustbox}
\end{table*}

We show classification accuracies.
Note that some methods in Tables \ref{tab:vs_others} and \ref{tab:vs_combi}, including VAT, performed image pre-processing with ZCA on CIFAR-10, which is not used in our experiments.
In terms of the model capacity, it is a fair comparison as all other methods used the same network architecture as ours.

In Table \ref{tab:vs_others}, we compared LVAT to VAT, other consistency regularizations, and also other approaches introduced in Section \ref{sec:relatedwork}.
We can see that LVAT substantially improved the original VAT,
and moreover, outperformed all of other methods in all eight experimental settings.

It has been reported that even better results can be obtained by combining consistency regularizations (MT, $\Pi$-Model, TempEns, and VAT) together with other techniques, such as graph-based method (SNTG).
We also compared our method to those combinations in Table \ref{tab:vs_combi}.
It is noteworthy that LVAT still surpassed all other methods, except for the result in the SSL testings on SVHN (VAT + Ent + SNTG) and on SVHN (w/ aug.) ($\Pi$-Model + SNTG).
In particular, LVAT-Glow on CIFAR-10 and CIFAR-10 (w/ aug.) showed outstanding performance.
We believe that combining LVAT with other methods (e.g., LVAT + SNTG) would also achieve further improved performance, but we leave it as future works.

\section{Discussions}
\subsection{Justification of Our Approach}
\label{sec:justification}
To justify our approach, we here confirmed the following two:
1) LVAT generates more adverse (i.e., effective) examples than VAT,
and 2) LVAT overcomes the local constraint of VAT.

\subsubsection{Generating more adverse examples.}
\begin{figure}[t]
	\centering
	\subfigure[Cross-entropy loss for test]{
		\includegraphics[width=0.22 \textwidth]{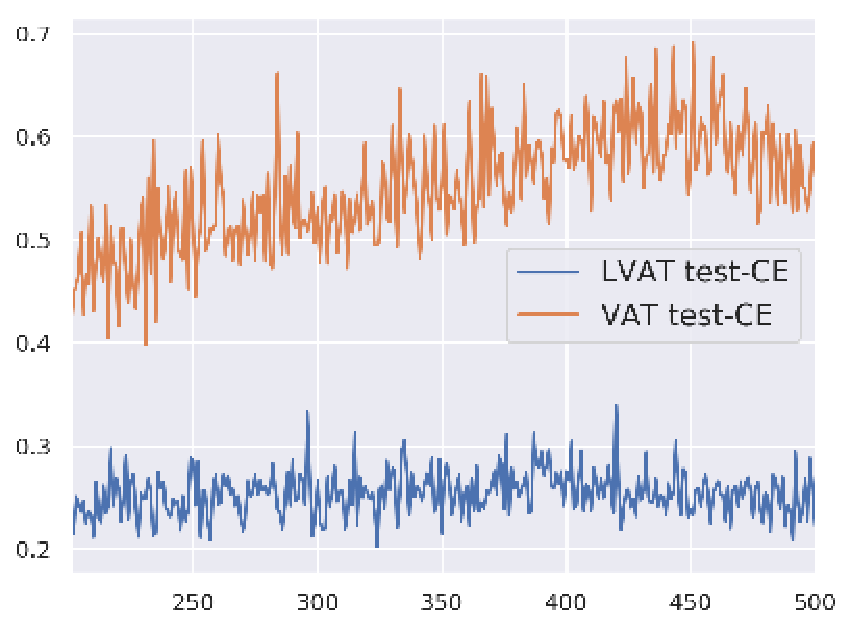}
		\label{fig:fig_ce_test}
	}
	\subfigure[Classification accuracy in test]{
		\includegraphics[width=0.22 \textwidth]{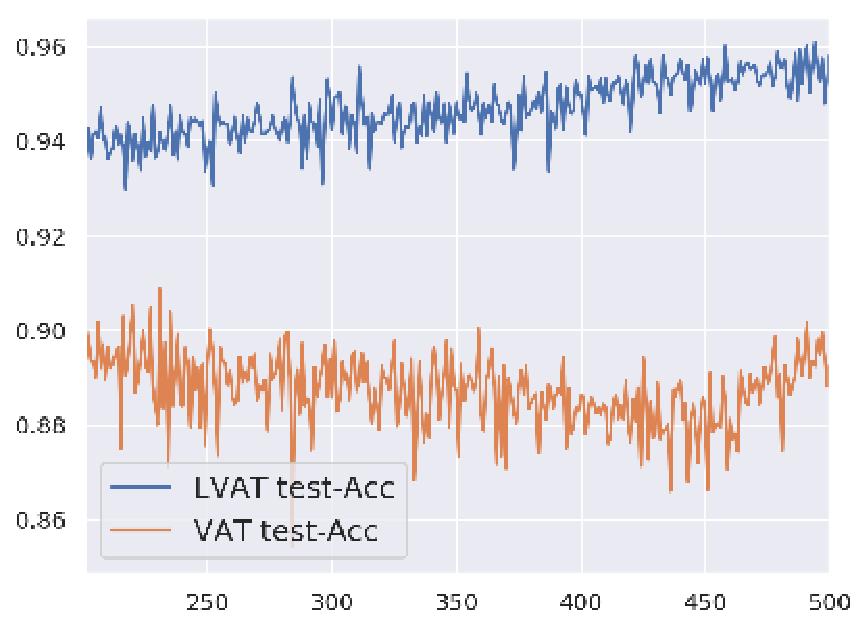}
		\label{fig:fig_acc_test}
	}
	\\
	\subfigure[Cross-entropy loss in training]{
		\includegraphics[width=0.22 \textwidth]{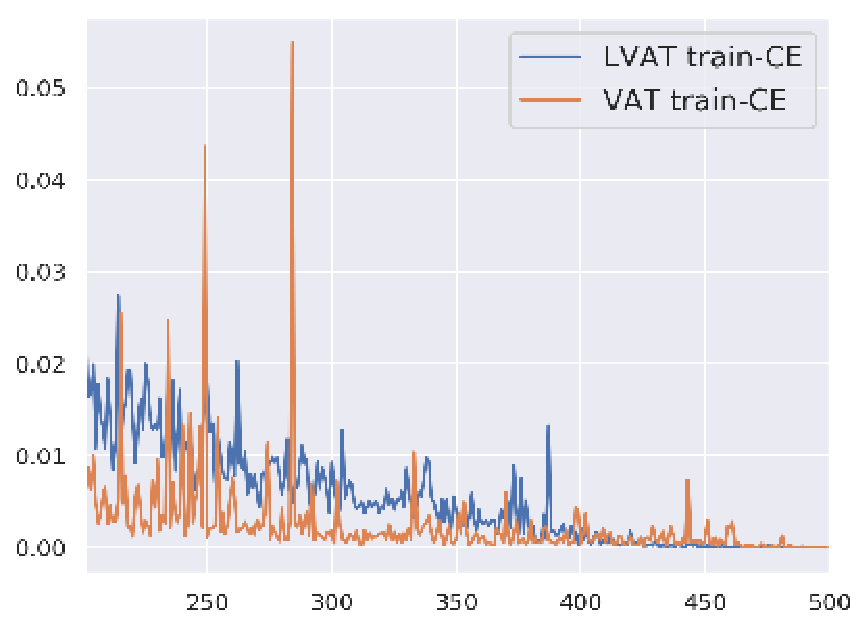}
		\label{fig:fig_ce_train}
	}
	\subfigure[Total loss in training]{
		\includegraphics[width=0.22 \textwidth]{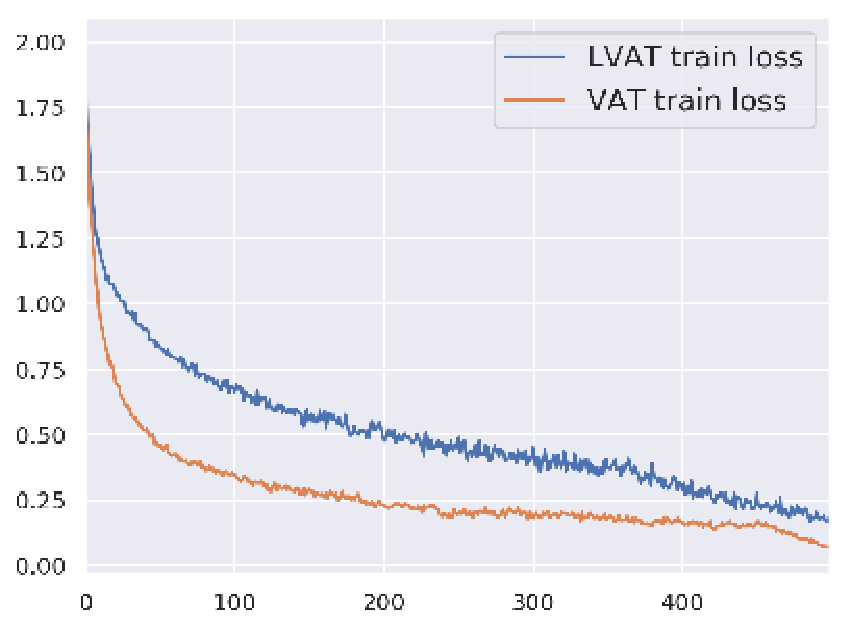}
		\label{fig:fig_loss_train}
	}
	\caption{Learning curve of classifier on CIFAR-10. 400 training iterations are denoted as 1 epoch.
	}
	\label{fig:loss}
\end{figure}
We perform two checks: one from the training process of the classifier, and the other from direct measurement.
First, we show the training process of the classifier on CIFAR-10 (SL) in Fig.\ \ref{fig:loss}, as measured in the experiments of the previous section.
In the figure, 400 training iterations are denoted as 1 epoch,
and the last 300 epochs of the 500-epoch are plotted in Figs.\ \ref{fig:fig_ce_test} to \ref{fig:fig_ce_train}.
Fig.\ \ref{fig:fig_ce_train} shows that the cross-entropy (CE) loss in the training is lower for VAT than for LVAT up to about 400 epochs.
However, for the test dataset, the CE loss for LVAT is consistently lower than VAT, with a correspondingly higher classification accuracy (Acc), as shown in Figs.\ \ref{fig:fig_ce_test} and \ref{fig:fig_acc_test}.
This indicates that LVAT yields higher generalization performance than VAT, in other words, LVAT works more powerfully than VAT as a regularizer.
The same is shown in Fig.\ \ref{fig:fig_loss_train}, which plots the total loss comprising the sum of the CE and the consistency loss, i.e., $L_{\text{lvat}}$ for LVAT or $L_{\text{vat}}$ for VAT.
Despite the fact that the CE loss of LVAT was only about 0.01 larger than that of VAT as we saw in Fig.\ \ref{fig:fig_ce_train}, Fig.\ \ref{fig:fig_loss_train} shows that the total loss of LVAT was at least 0.1 greater than that of VAT throughout, indicating that $L_{\text{lvat}}$ was much higher than $L_{\text{vat}}$.
Thus it shows that the regularization by LVAT is more powerful than that by VAT, meaning that LVAT is able to generate more powerful adversarial examples than VAT.

\begin{figure}[t]
	\centering
	\subfigure[SVHN]{
		\includegraphics[width=0.22 \textwidth]{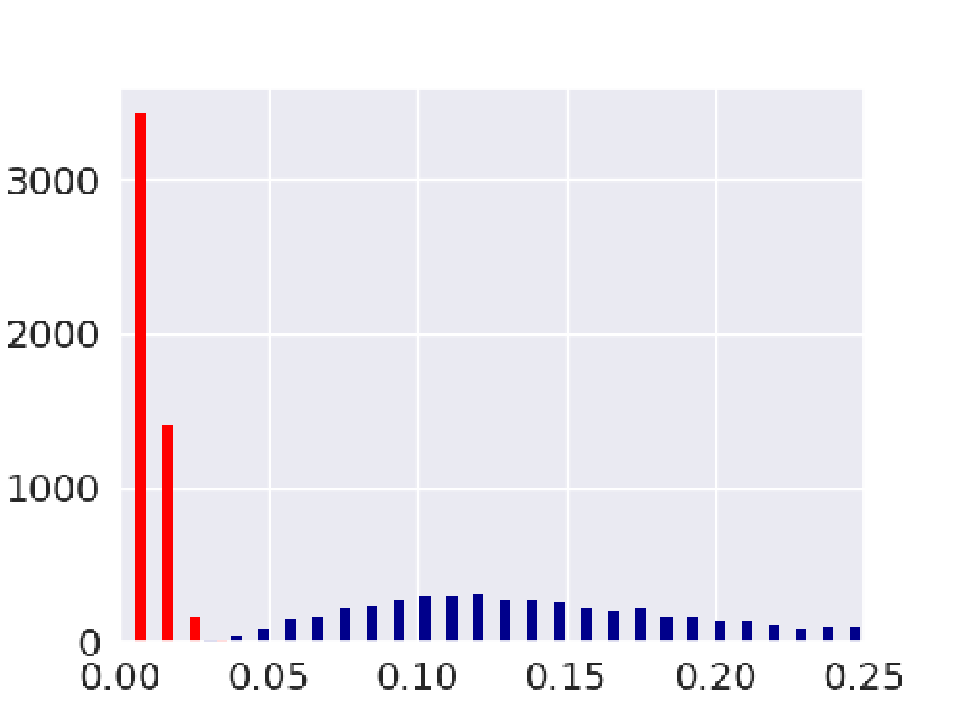}
	}
	\subfigure[CIFAR-10]{
		\includegraphics[width=0.22 \textwidth]{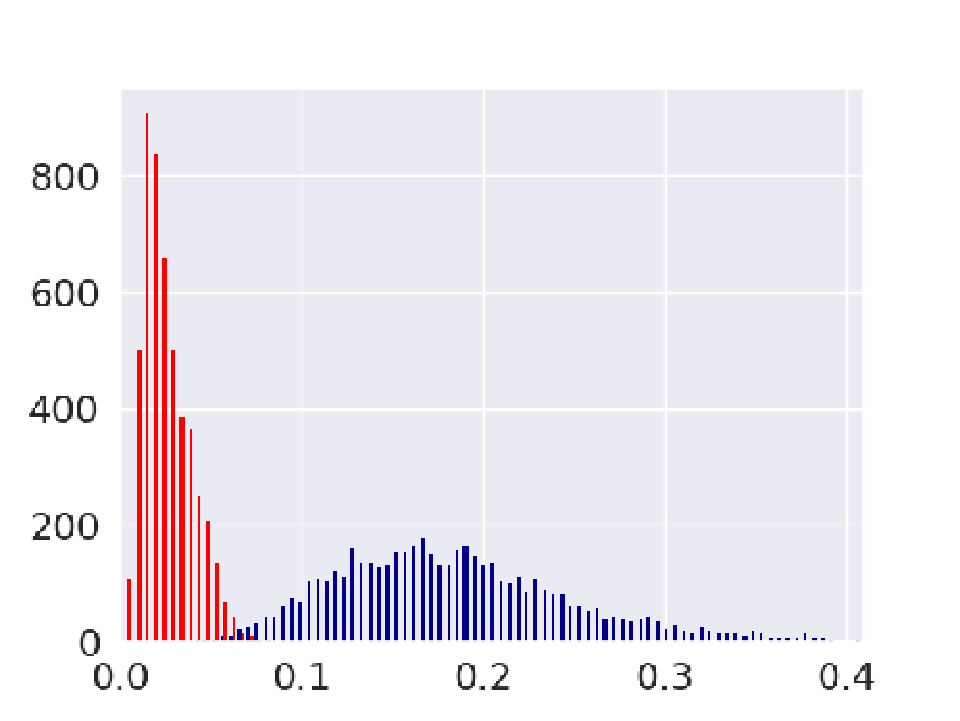}
	}
	\caption{Histograms of consistency cost. x-axis is consistency cost as KL divergence and y-axis is frequency.
		Blue is $L_{\text{lvat}}$ for LVAT-Glow and red is $L_{\text{vat}}$.
		For both datasets, LVAT generates more adverse (i.e., effective) samples than VAT.
	}
	\label{fig:kl}
\end{figure}


We also directly compared LVAT with VAT in terms of the consistency cost, which indicates how adverse the samples the methods generate are.
Specifically, we compared $L_{\text{lvat}}$ in Eq.\ (\ref{eq:loss_lvat}) with $L_{\text{vat}}$ in Eq.\ (\ref{eq:loss_vat}), both of which are measured in KL divergence between the model predictions for the input with and without perturbation.
We used 5000 samples selected at random from the test sets for both datasets.
$L_{\text{lvat}}$ and $L_{\text{vat}}$ were computed using the classifiers trained 1 epoch with LVAT-Glow and VAT, respectively.
Fig.\ \ref{fig:kl} shows the histograms, and it turned out that distributions of $L_{\text{lvat}}$ (blue) were obviously larger than those of $L_{\text{vat}}$ (red) on both datasets
From this result, we can see that LVAT generates more adverse samples than VAT.

\subsubsection{Overcoming local constraint.}
\begin{figure}[t]
	\centering
	\subfigure[LVAT-VAE]{
		\includegraphics[width=0.22 \textwidth]{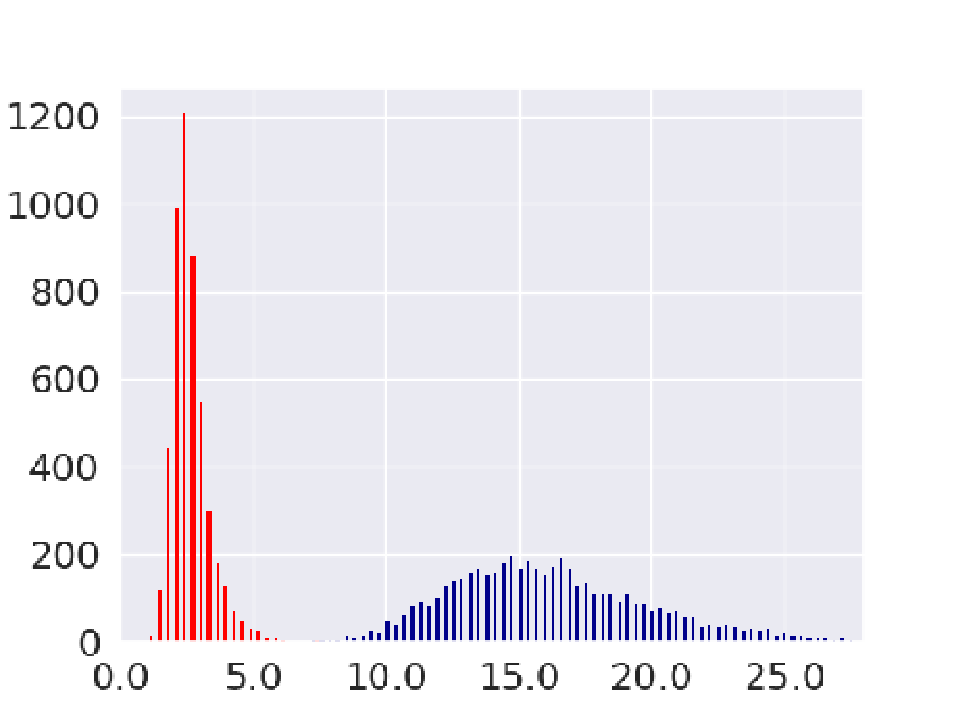}
		\label{fig:diff_w_vae}
	}
	\subfigure[LVAT-Glow]{
		\includegraphics[width=0.22 \textwidth]{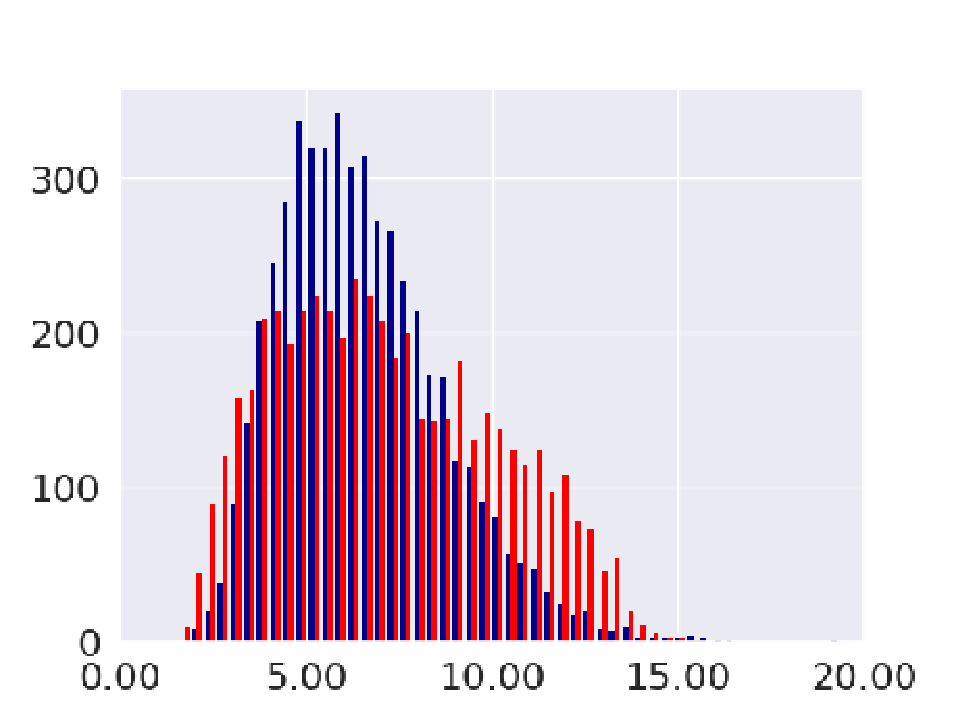}
		\label{fig:diff_w_glow}
	}
	\caption{Histograms of $L_{2}$ distance between original images and adversely perturbed ones.
		x-axis is $\lVert {\bf x} - \text{Dec} (\text{Enc}({\bf x})  + {\bf r}_{\text{lvat}})\rVert_{2}$, and y-axis is frequency.
		Red is SVHN and blue is CIFAR-10, and for both $\epsilon_{\text{lvat}} = 1.0$.
		For each dataset 5,000 samples are randomly sampled.
		This indicates that LVAT generates perturbations of various magnitudes without local constraint.
	}
	\label{fig:ep_LVAT}
\end{figure}
If LVAT successfully overcame the local constraint, LVAT should be able to generate adversarial examples with various magnitudes.
We thus visualized the magnitude of perturbation observed in the input space.
Specifically, we measured $L_{2}$ distance between the original input images and the adversarial images that LVAT generates, i.e., $\lVert {\bf x} - \text{Dec} (\text{Enc}({\bf x})  + {\bf r}_{\text{lvat}})\rVert_{2}$.
Fig.\ \ref{fig:ep_LVAT} shows the histograms with the same samples and classifier as the above.
It is shown that LVAT generates adversarial examples in the wide range of magnitude, unlike in the original VAT where every adversarial example is generated with the same given magnitude $\epsilon_{\text{vat}}$.
Thus we can see that LVAT overcomes the local constraint and can generate adversarial examples flexibly.
Also we can see that while the distributions of the two datasets (i.e., blue and red) overlap in LVAT-Glow,
they are completely separated in LVAT-VAE.
Indeed, we consider that it causes LVAT-VAE's low performance on CIFAR-10 and will discuss this later.


\subsection{Visual appearance of adversarial examples.}
\begin{figure}[t]
	\centering
	\subfigure[LVAT-VAE on SVHN]{
		\includegraphics[width=0.47 \textwidth]{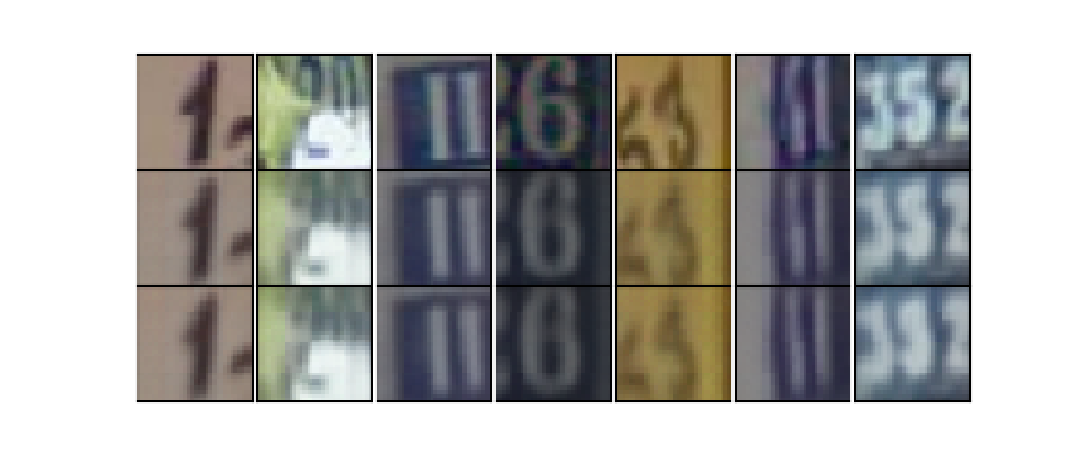}
		\label{fig:svhn_lvat_vae}
	}
	\subfigure[LVAT-VAE on CIFAR-10]{
		\includegraphics[width=0.47 \textwidth]{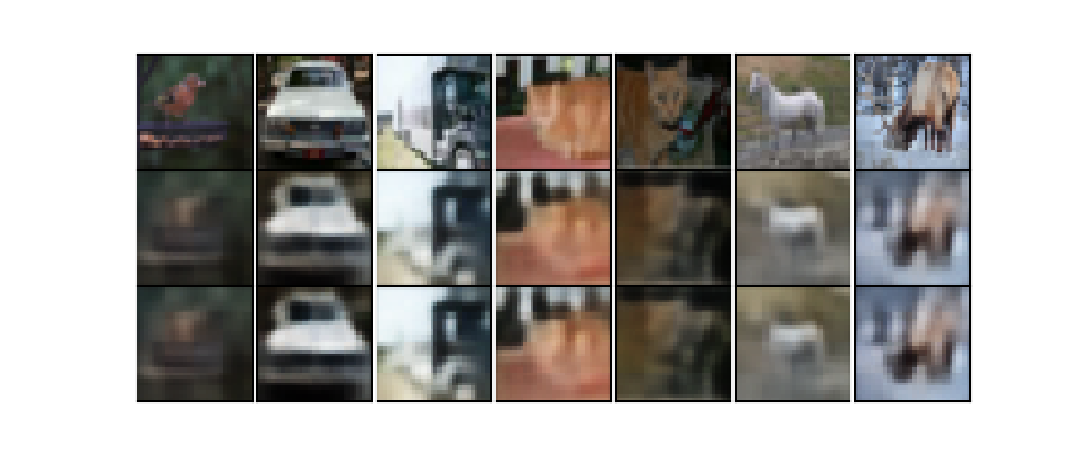}
		\label{fig:cifar10_lvat_vae}
	}
	\\
	\subfigure[LVAT-Glow on SVHN]{
		\includegraphics[width=0.47 \textwidth]{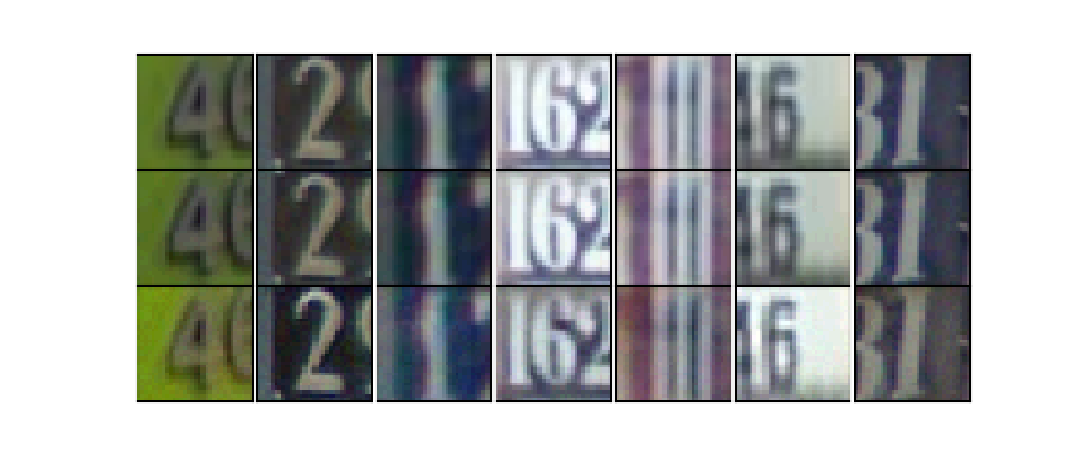}
		\label{fig:svhn_lvat_glow}
	}
	\subfigure[LVAT-Glow on CIFAR-10]{
		\includegraphics[width=0.47 \textwidth]{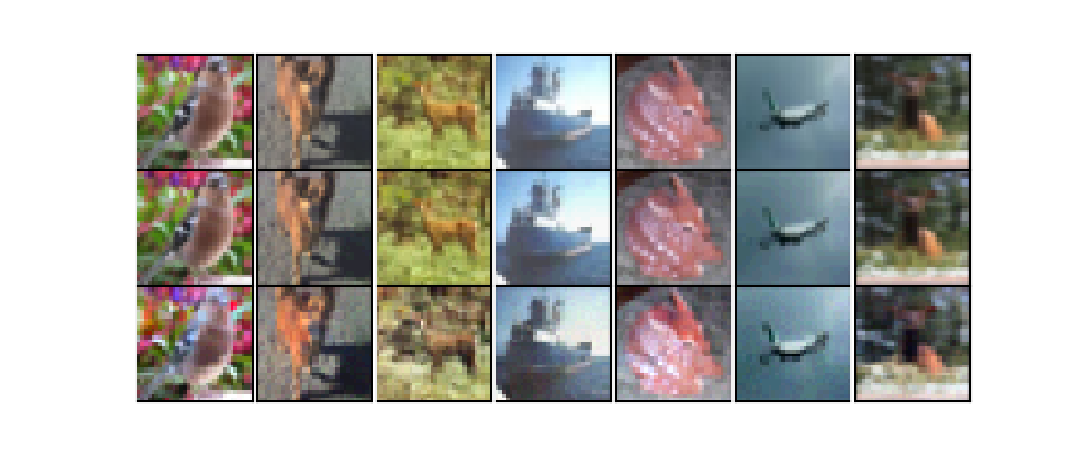}
		\label{fig:cifar10_lvat_glow}
	}
	\\
	\subfigure[VAT on SVHN]{
		\includegraphics[width=0.47 \textwidth]{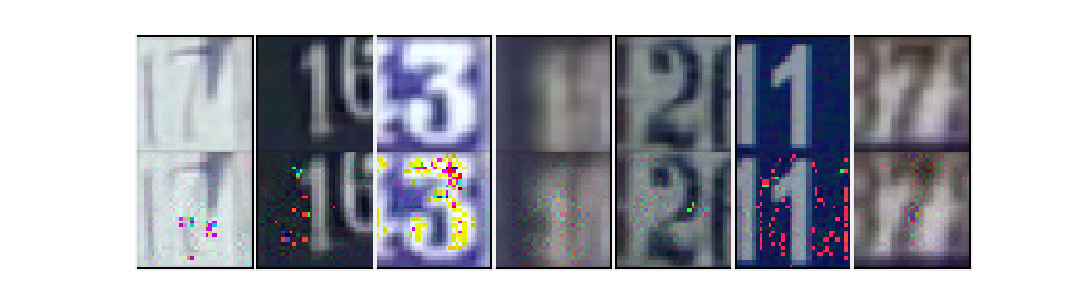}
		\label{fig:svhn_vat}
	}
	\subfigure[VAT on CIFAR-10]{
		\includegraphics[width=0.47 \textwidth]{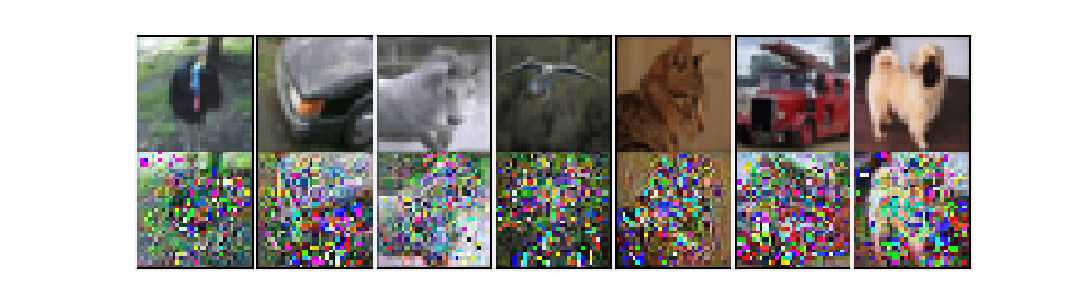}
		\label{fig:cifar10_vat}
	}
	\caption{Generated Images.
		For (a) through (d), first row: original images ${\bf x}$, second row: reconstructed images via generative model without perturbation $ \hat{{\bf x}} = \text{Dec} (\text{Enc}({\bf x}))$, third row: adversarial images ${\bf x}_{\text{adv}} = \text{Dec} (\text{Enc}({\bf x})  + {\bf r}_{\text{lvat}})$.
		For (e) and (f), first row: ${\bf x}$, second row: ${\bf x}_{\text{adv}} = {\bf x} + {\bf r}_{\text{vat}}$.
	}
	\label{fig:imgs}
\end{figure}
Next, we see the visual appearance of adversarial examples of LVAT and VAT in Fig.\ \ref{fig:imgs}.
We note that the adversarial samples in Fig.\ \ref{fig:imgs} were created with the same
perturbation magnitudes, $\epsilon_{\text{lvat}}$ and $\epsilon_{\text{vat}}$, as the ones we used for the performance evaluation (see Section \ref{sec:param}).
It means that by means of the consistency cost, the classifier has been trained so that it outputs the same predictions for the images in the bottom rows and the ones in the top rows.
We can see that adversarial images of VAT are tainted with artifacts, whereas the ones of LVAT look realistic.
In both generative models VAE and Glow, the latent space $p({\bf z})$ is constructed so that the points in the high-density area in $p({\bf z})$ correspond to the data used during the model training, i.e., real images.
Thus, unless $\epsilon_{\text{lvat}}$ is not too large, the perturbed latent representation ${\bf z}_{\text{adv}}$ computed in LVAT still should correspond to a realistic image.
In VAT, on the other hand, there is no such thing to guide the quality of the generated images.
As described in Section \ref{sec:problem}, even when $\epsilon_{\text{vat}}$ is too large for an images , VAT has no way to adjust and limit the magnitude to avoid generating artifacts.
Also, it has been argued in \cite{xie2019unsupervised} that these noisy images generated in VAT seem harmful to further performance improvement.
Unlike VAT, there is no such concern in LVAT.

\subsection{Failure analysis: limitation of VAE reconstruction ability on CIFAR-10}
\label{sec:vae_ability}

\begin{figure}[t]
	\centering
	\subfigure[VAE]{
		\includegraphics[width=0.3 \textwidth]{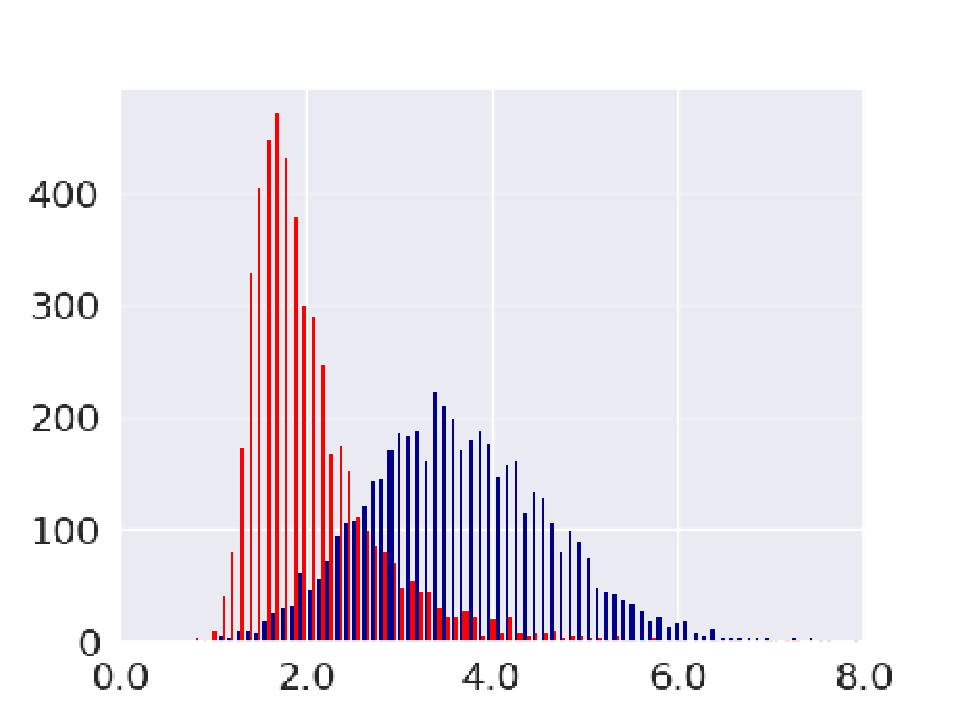}
		\label{fig:diff_wo_vae}
	}
	\subfigure[Glow]{
		\includegraphics[width=0.3 \textwidth]{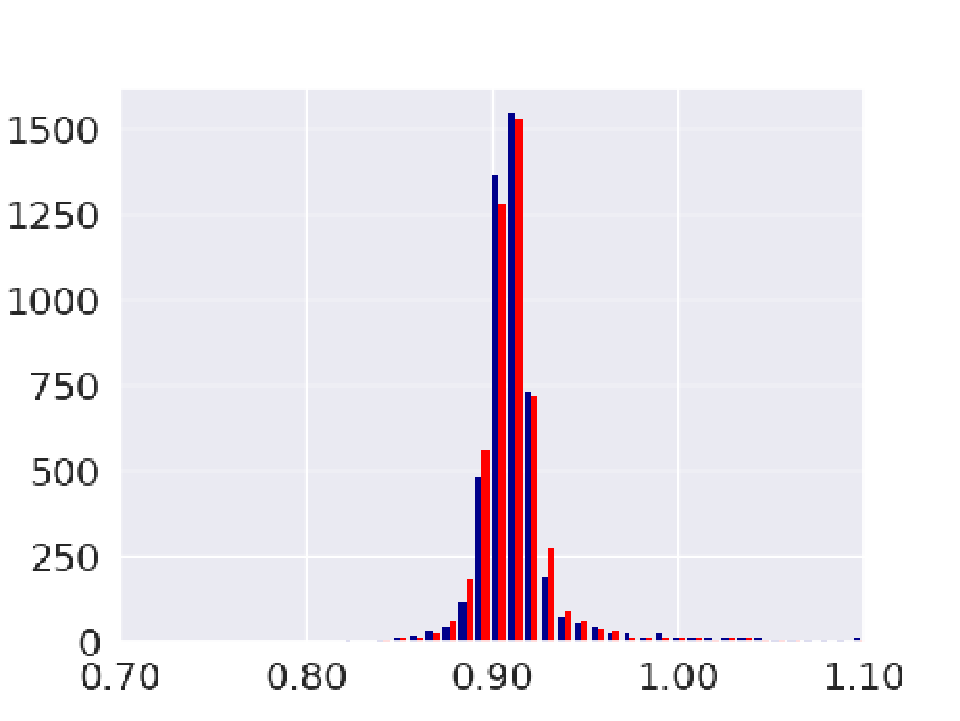}
		\label{fig:diff_wo_glow}
	}
	\caption{Histograms of reconstruction error.
		x-axis is $\lVert {\bf x} - \text{Dec} (\text{Enc}({\bf x}))\rVert_{2}$ and y-axis is frequency.
		This figure corresponds to Fig.\ \ref{fig:ep_LVAT}.
		Red is SVHN and blue is CIFAR-10, and for both $\epsilon_{\text{lvat}} = 1.0$.
		This indicates that regardless of perturbation, reconstruction error of VAE is larger on CIFAR-10 than on SVHN.
	}
	\label{fig:diff}
\end{figure}
Our proposed method LVAT achieved good results as we saw, especially LVAT-Glow on CIFAR-10.
However, it also turned out that the error rates of LVAT-VAE on CIFAR-10 were higher than the other experimental settings.
We analyze its reason in this section.
We found that on CIFAR-10, while the adversarial images with Glow are very sharp, those with VAE are blurred, as shown in the bottom rows in Figs.\ \ref{fig:cifar10_lvat_glow} and \ref{fig:cifar10_lvat_vae}.
More importantly,  we can see that the images just reconstructed without perturbation are also blurry with VAE on CIFAR-10 (the middle row in Fig.\ \ref{fig:cifar10_lvat_vae}).
It shows that regardless of perturbation, just passing Enc() and Dec() of VAE will blur the input image, which is a known characteristic of VAE \cite{chen2016variational}.
To examine this statistically, we measured the reconstruction error for the two datasets.
That is, $L_{2}$ distance between the original images and the decoded images, $\lVert {\bf x} - \text{Dec} (\text{Enc}({\bf x}))\rVert_{2}$, and we compared VAE and Glow about it.
Fig.\ \ref{fig:diff} shows the histograms, and we can see the following two things from them.
First one is that the reconstruction error in VAE is apparently larger than that in Glow.
Second one, which is more important, is that VAE caused much larger reconstruction errors on CIFAR-10 (blue) than those on SVHN (red) (Fig.\ \ref{fig:diff_wo_vae}),
while Glow (Fig.\ \ref{fig:diff_wo_glow}) did not show difference between the datasets.
These statistics are consistent with what we saw visually in Fig.\ \ref{fig:imgs}.
We attributed the larger reconstruction error of VAE on CIFAR-10 compared to those on SVHN to the complexity of images contained in the datasets.
Given these observations and the fact that the classification performance of LVAT-Glow was very good on CIFAR-10, we conclude that
the reconstruction ability of the generative model is crucial to the quality of regularization, which caused the high error rates of LVAT-VAE on CIFAR-10.


\section{Conclusion}
We focused on the local constraint of VAT: VAT can generate adversarial perturbation only within a very small area around the input data point with fixed magnitude.
In order to circumvent this constraint, we proposed LVAT in which computing and injecting perturbation are done in the latent space.
Since adversarial examples in LVAT are generated via the latent space, they are more flexible than those in the original VAT, which led to more effective consistency regularization and better classification performance as a result.
We compared LVAT with VAT and other state-of-the-art methods in supervised and semi-supervised scenarios for a classification task in SVHN and CIFAR-10 datasets (both with and without data-augmentation).
Our evaluation indicates that LVAT outperforms state-of-the-art methods in terms of classification accuracy in different scenarios.

\section*{Acknowledgement}
This work was supported in part by JSPS KAKENHI Grant Number 20K11807.

\bibliographystyle{splncs04}
\bibliography{my_bib_210321}

\end{document}